\documentclass[a4paper]{article}

\usepackage[english]{babel}
\usepackage[utf8x]{inputenc}
\usepackage[T1]{fontenc}

\usepackage[a4paper,top=3cm,bottom=2cm,left=3cm,right=3cm,marginparwidth=1.75cm]{geometry}

\usepackage{amsmath}
\usepackage{graphicx}
\usepackage[colorinlistoftodos]{todonotes}
\usepackage[colorlinks=true, allcolors=blue]{hyperref}

\usepackage{lineno,hyperref}
\usepackage{subfigure}
\usepackage{algorithmic}
\usepackage{float}
\usepackage[affil-it]{authblk}

\title{Solving Mixed Model Workplace Time-dependent Assembly Line Balancing Problem with FSS Algorithm}
\author[1]{J.B. Monteiro-Filho*}
\author[1]{I.M.C. Albuquerque}
\author[1]{F.B. Lima Neto}
\affil[1]{Computational intelligence research group - Polytechnical School of Pernambuco Benfica 455, Recife-PE, Brazil}
\affil[*]{Corresponding author: jbmf@ecomp.poli.br}

\begin{document}
\maketitle

\begin{abstract}
Balancing assembly lines, a family of optimization problems commonly known as Assembly Line Balancing Problem, is notoriously NP-Hard. They comprise a set of problems of enormous practical interest to manufacturing industry due to the relevant frequency of this type of production paradigm. For this reason, many researchers on Computational Intelligence and Industrial Engineering have been conceiving algorithms for tackling different versions of assembly line balancing problems utilizing different methodologies. In this article, it was proposed a problem version referred as Mixed Model Workplace Time-dependent Assembly Line Balancing Problem with the intention of including pressing issues of real assembly lines in the optimization problem, to which four versions were conceived. Heuristic search procedures were used, namely two Swarm Intelligence algorithms from the Fish School Search family: the original version, named ``vanilla", and a special variation including a stagnation avoidance routine. Either approaches solved the newly posed problem achieving good results when compared to Particle Swarm Optimization algorithm.
\end{abstract}

\section{Introduction}
\label{intro}

When Henry Ford created Assembly Lines (AL) back in 1913, he dramatically and indelibly modified  manufacturing industries production paradigm \cite{tu2001impact}. Because of its smashing success, the use of AL has been continuously spreading over the world and most of manufacturing industries still use it greatly.

Formally, an assembly line is a flow based production system in which the work units, denominated workstations, are disposed in a series manner \cite{Boysen2008}.  The work pieces travel along the line are moved by a transport system. There are many elements related to assembly lines productivity such as the distribution of the assembly tasks on each workstation. This problem is commonly referred in literature as Assembly Line Balancing Problem (ALBP). 

Many different exact and heuristic procedures have been developed to tackle planning and schedulling problems \cite{yang2016multi,mencia2016genetic,kyriklidis2016evolutionary,jia2014multiobjective}, and specifficaly the assembly line related problems \cite{Scholl2006}. Furthermore, metaheuristic approaches have been playing an important role in the real life engineering \cite{boulkaibet2015finite} optimization tasks such as combinatorial optimization trying to solve them in realistic time \cite{blum2003metaheuristics}. Recently, a metaheuristic algorithm called Fish School Search (FSS) was modified and applied to solve a simple version of an assembly line balancing problem \cite{albuquerque2016, de2016fish}. In those works, a mapping procedure was utilized in order to perform the search in a continuous space and to convert the fishes position (\textit{i.e.} candidate solutions) into a discrete sequence of tasks only when it was necessary to calculate fitnesses for the fishes.

The present work intends to apply the aforementioned mapping approach in a novel version of the assembly line balancing problem called Mixed Model Workplace Time Dependent Assembly Line Balancing Problem (MMWALBP). This version of the problem takes into account some features of assembly lines of complex and large sized products such as cars an trucks.

The remainder of this article is organized as follows: Section \ref{MMWALBP} defines the version of the assembly line balancing problem proposed, Section \ref{FSS} explain the two used versions of the Fish Schol Search algorithm referred as FSS-Vanilla and a variation of it denominated FSS-SAR, presented originally in the works of Monteiro \textit{et al.} \cite{monteiro2016, monteiro2016improved}. Section \ref{MMWALBPFSS} presents the solution approach utilized to tackle MMWALBP with FSS-V and FSS-SAR and the results obtained are shown and commented upon in Sections \ref{experiments} and \ref{conclusion}, respectively.

\section{Mixed Model Workplace Time Dependent Assembly Line Balancing Problem - MMWALBP}
\label{MMWALBP}

\subsection{New problem definition}
\label{problemdefinition}

Many different versions of assembly line balancing problems have been conceived by researchers aiming to include in their analyses the different features of real assembly lines. Some of these are:

\begin{enumerate}

\item Mixed Model Assembly Line Balancing Problem (MMALBP): This version of ALBP treats the flexible assembly lines in which setup times between models are irrelevant for variations of a basis product version. There exists a difference between mixed-model and mixed-product assembly lines. In the first case, the production sequence is composed of a mixed ordering of the different versions of the same model. In the second case, the production is performed in long sequence batches of each model \cite{Thomopoulos1967}.
\item Two Sided Assembly Line Balancing Problems (TALBP): In this type of assembly lines, there are two workplaces per workstation named \textit{Right} and \textit{Left} workplaces. Tasks are classified as \textit{Right}/\textit{Left}, \textit{i.e}. the tasks which have to be necessarily executed on the right/left side of the assembly line, or \textit{Either}, \textit{i.e.} the tasks that can be performed on either sides of the line \cite{bartholdi1993balancing}.
\item Mixed Model Two Sided Assembly Line Balancing Problem (MMTALBP): Simultaneous mixed model and two sided assembly line balancing problem \cite{Simaria2009a}.
\item Multi Manned Assembly Line Balancing Problem (MALBP): The multi manned assembly lines allow several workplaces in each workstation \cite{Dimitriadis2006}.
\item Variable Workplace Assembly Line Balancing Problem (VWALBP): An ALBP is known to be a VWALBP when different workplaces are set in every workstation, and also, every task has a definition of which workplace it should be performed at \cite{Becker2009}.

\end{enumerate}

In this work, a novel version of ALBP was proposed taking simultaneously into account the features present in the versions represented by the items 1, 2, 4 and 5: Mixed models and many workplaces per workstation. This version is referred as Mixed Model Workplace Time-dependent Assembly Line Balancing Problem (MMWALBP). More specifically, it was found necessary to include the two features described next in a single problem version in order to make it utilizable mainly for assembly lines producing big sized products:

\paragraph{Mixed model nature:} In a single model assembly line, a high volume of standardized homogeneous products are produced. This is not suitable for customers demand with high variety. In the other hand, mixed model assembly lines are widely used to improve the flexibility to adapt to changes in market demand in a range of industries. In mixed-model assembly lines, two or more products with similar production characteristics are produced on the same assembly line \cite{Delice2014}.

In mixed model balancing problems, many authors utilize the mean model approach, first introduced in the work of Thomopoulos \cite{Thomopoulos1967}. The same approach was utilized in MMWALBP. The mean model is utilized to convert a mixed model problem in a single model one, creating a virtual model for which the task times are adjusted based on the production plan for each model. Further, a joint precedence graph has to be built.

\paragraph{Many workplaces per workstation:} Multi-Manned and Two Sided Assembly Lines Balancing Problems, referred in this work as MALBP and TALBP, try to include the allowance of many workplaces into a single workstation. This feature is common in many assembly lines, mainly those manufacturing big sized products such as cars and trucks. However, the two mentioned versions treat this feature in different ways. MALBP gives the decision maker the possibility to set the maximum number of workplaces per workstation. However, there is no specification of which workplace should execute a specific task. In TALBP, only two workplaces are allowed in a single workstation, commonly named \textit{Right} and \textit{Left} workplaces. The decision maker cannot set more workplaces in the same workstation. Differently of MALBP, in TALBP the tasks can only be assigned to a specific workplace, which means that every task has an extra attribute that is valued \textit{Right}/\textit{Left} when the task can only be assigned to \textit{Right}/\textit{Left} workplaces, and \textit{Either} when it occurs that some task can be assigned to either \textit{Right} or \textit{Left} workplaces. Hence, these two different problem versions try differently to model the same real line feature, which is more than one workplace per workstation. The main goal of using multi-manned workstations, \textit{i.e.} workstations containing more than one workplace, is to minimize the number of workstations of the line while its total effectiveness (in terms of number of workers) remains optimal \cite{Roshani2013}. Multi-manned assembly lines have substantial advantage over a simple assembly line such as reducing the length of the assembly line and consequently the Work in Progress number of products, Total Throughput Time, cost of tools and fixtures, material handling, workers movement and setup times  \cite{Fattahi2011}.\\

The two main features of MALBP and TALBP were combined originating the Variable Workplace Assembly Line Balancing Problem (VWALBP) \cite{Becker2009}. In VWALBP, the workstation is divided in 8 possible workplaces corresponding to eight different zones within the workstation. Each task contains an attribute that tells the decision maker in which zone it should be performed. A maximum number of open/active workplaces (operators) per workstation is defined. The assignment is performed taking into account a list of prohibited workplaces combinations. This list contains the pair of workplaces for which a displacement from one to the other would take a relevant amount of time. This means that if the pair of workplaces \textit{a} and \textit{b} is prohibited, if a task from \textit{a} is assigned to an operator, thus no task from \textit{b} could be assigned to the same operator.

In this work, the defined problem version MMWALBP has a similar approach when compared to VWALBP, but the difference relies on the fact that there is no need for a list of prohibited workplaces association. Instead, a time correction procedure was utilized in a way that every displacement between different zones within the workstation generates an additional time in the task that required the displacement. Times correction procedure and also idle times controlling will be further detailed in the following sections.

\subsection{Proposed Controlling Mechanisms for MMWALBP}
\label{controlMechanisms}

\subsubsection{Times Correction}
\label{timescorrection}

\begin{figure}[htpb]
\centering
\caption[Zones definition]{Zones definition}
\label{workplaces}
\includegraphics[width=7cm,height=4cm]{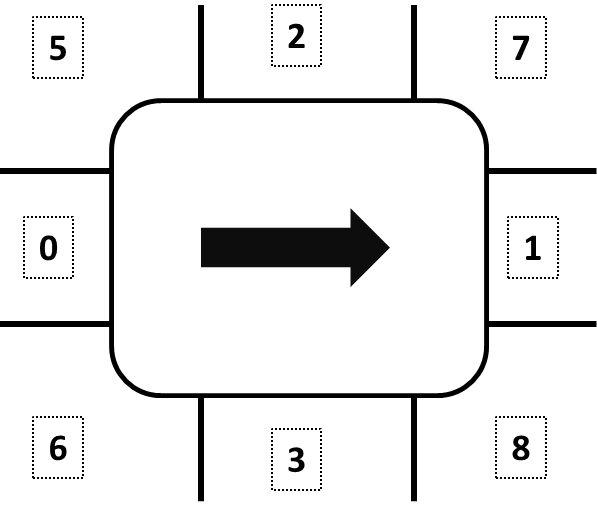}
\end{figure}

Similarly to what has been done in VWALBP, the workstation was divided in 8 different zones as it is shown in Figure \ref{workplaces}. The arrow indicates the flow sense of the line. It can be seen that each number identify a work zone within the workstation that will be used to define workplaces. Number 4 represents the zone in the interior of the product to be assembled.

A displacement matrix is applied to define the time cost in order for an operator to move from on work zone to another. An example of a time increment matrix is shown in Table \ref{timesMatrixExample}. This time matrix should be used to input the displacement times in the operations when some work zone change is needed. The rows and columns represent the origin and destination work zones, respectively. The displacement times matrix is specific for each assembly line, hence it should be specifically defined for every line before tackling its Assembly Balancing Problem. Many different approaches exist in order to define human operations standard times. One of most common and used approaches is the Method-Times Measurement (MTM) \cite{kanawaty1992introduction}\cite{longo2009effective}.

\begin{table}[]
\centering
\caption{Times correction matrix example}
\label{timesMatrixExample}
\begin{tabular}{cccccccccc}
& \textit{\textbf{0}} & \textit{\textbf{1}} & \textit{\textbf{2}} & \textit{\textbf{3}} & \textit{\textbf{4}} & \textit{\textbf{5}} & \textit{\textbf{6}} & \textit{\textbf{7}} & \textit{\textbf{8}} \\
\textit{\textbf{0}} & 0                   & 0.06                & 0.03                & 0.03                & 0                   & 0.015               & 0.015               & 0.045               & 0.045               \\
\textit{\textbf{1}} & 0.06                & 0                   & 0.03                & 0.03                & 0                   & 0.045               & 0.045               & 0.015               & 0.015               \\
\textit{\textbf{2}} & 0.03                & 0.03                & 0                   & 0.06                & 0                   & 0.015               & 0.045               & 0.015               & 0.045               \\
\textit{\textbf{3}} & 0.03                & 0.03                & 0.06                & 0                   & 0                   & 0.045               & 0.015               & 0.045               & 0.015               \\
\textit{\textbf{4}} & 0                   & 0                   & 0                   & 0                   & 0                   & 0                   & 0                   & 0                   & 0                   \\
\textit{\textbf{5}} & 0.015               & 0.045               & 0.015               & 0.045               & 0                   & 0                   & 0.03                & 0.03                & 0.06                \\
\textit{\textbf{6}} & 0.015               & 0.045               & 0.045               & 0.015               & 0                   & 0.03                & 0                   & 0.06                & 0.03                \\
\textit{\textbf{7}} & 0.045               & 0.015               & 0.015               & 0.045               & 0                   & 0.03                & 0.06                & 0                   & 0.03                \\
\textit{\textbf{8}} & 0.045               & 0.015               & 0.045               & 0.015               & 0                   & 0.06                & 0.03                & 0.03                & 0                  
\end{tabular}
\end{table}

\subsubsection{Precedence Relations and Idle Times Controlling}
\label{precedenceidleness}

In all versions of assembly lines balancing problems that allow more than one workplace per workstation, precedence relations between tasks assigned to different operators within the same workstations have to be considered. This means that, if task $a$ precedes task $b$, it is necessary to guarantee that the start time of task $b$ is higher than the end time of task $a$.

The same approach used in the work of Simaria \textit{et al.} \cite{Simaria2009a} was applied in MMWALBP to keep track of the start and end times of each task. This means that every time a task is assigned to a specific operator, all the tasks assigned to the other operators within the same workstation must be checked. If any task in a different workplace precedes the current task and has its end time higher than the start time of the task to be allocated, the start time has to be set to be equal to the end time of its precedent.

\begin{figure}[htpb]
\centering
\caption[Idle times definition]{Idle times definition}
\label{idleTimes}
\includegraphics[width=8cm,height=5cm]{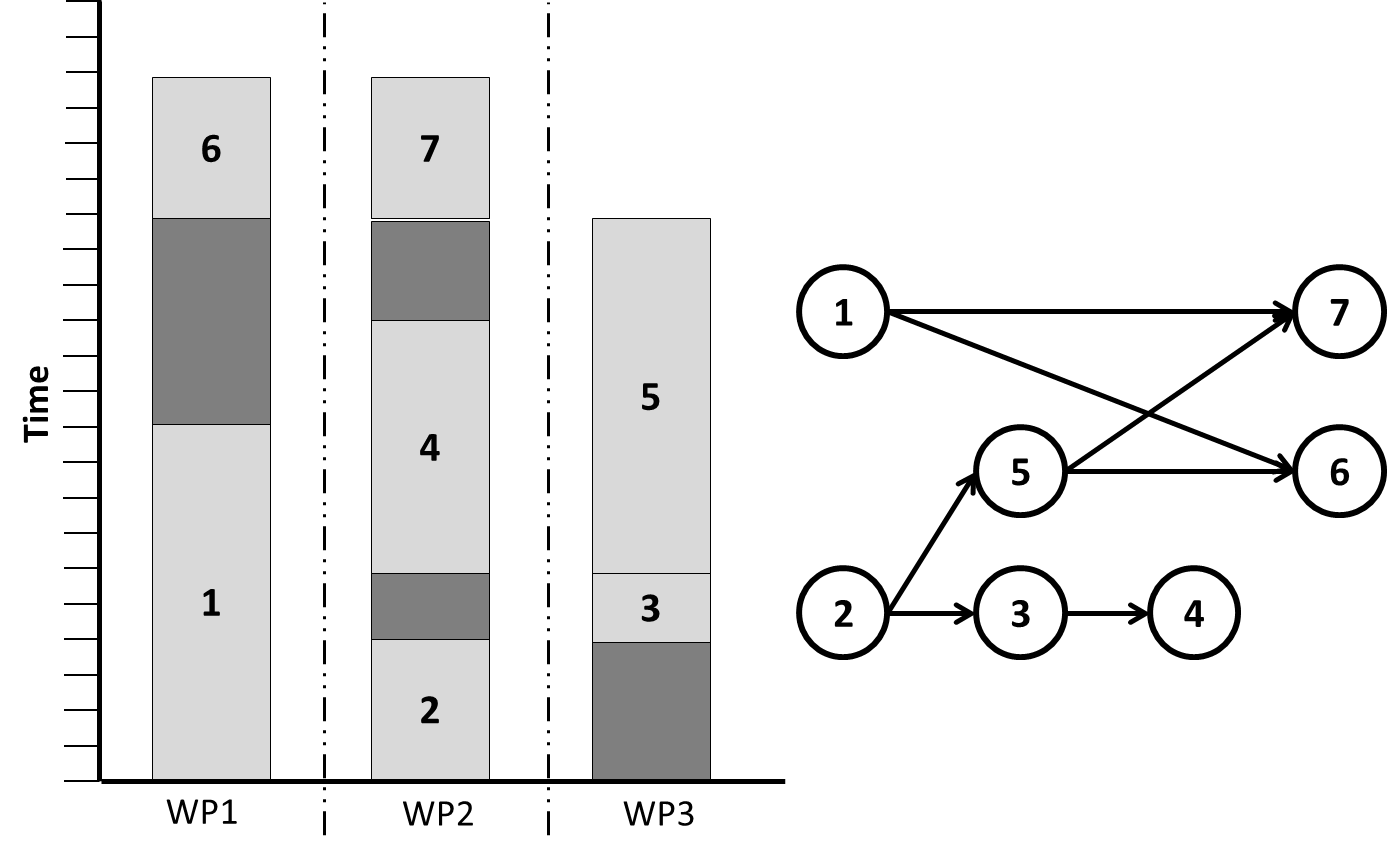}
\end{figure}

This procedure can be further explained with the assignment example showed in Figure \ref{idleTimes}. From the precedence graph, it can be seen that task 2 precedes task 3. For this reason, the start time of task 3 in the workplace 3 was set to be equal to the end time of task 2 in workplace 2. The same happens when task 4 is to be allocated. Task 3 must be finished before task 4 starts. The start time of task 4 is set to be equal to the end time of task 3. The same happens again for the tasks 6 and 7 which had their start times set to be equal to the end time of task 5. The dark gray bars indicate the presence of idle times within the workplace. In that period, the operator waits the finishing of a precedent task in a parallel workplace.

\section{Fish School Search Algorithm}
\label{FSS}

\subsection{FSS - Vanilla Version}
\label{FSSV}

Fish School Search (FSS) \cite{Filho2008} is a population based search algorithm inspired in the behavior of swimming fishes in schools that expand and contract while searching for food. This algorithm is inspired in the behavior of swimming fishes in schools that expand and contract while searching for food. Each fish position in the aquarium is a n-dimensional representation of a possible solution for the optimization problem. The algorithm makes use of weights for all the fishes for indicating cumulative success in the search for each fish in the school.

FSS is composed of the feeding and movement operators, the latter being divided into three sub-components, which are:

\begin{enumerate}

\item \textbf{Individual component of the movement operator:} every fish in the school performs a local search looking for promising regions in the search space. It is carried out as indicated in Eq. \ref{indmov}:

\begin{equation}
\label{indmov}
\textbf{x}_i(t+1)=\textbf{x}_{i}(t)+\textbf{rand}(-1,1)step_{ind},
\end{equation}

where $\textbf{x}_{i}(t)$ and $\textbf{x}_{i}(t+1)$ represent the position of the fish $i$ before and after the individual moment operator, respectively. And $\textbf{rand}(-1,1)$ is a uniformly distributed random number varying from $-1$ up to $1$ and $step_{ind}$ is a parameter that defines the maximum displacement for this movement. The new position $\textbf{x}_i(t+1)$ is only accepted if the fitness of the fish improves with the position change. If it is not the case, the fish remains in the same position for that round and $\textbf{x}_i(t+1)=\textbf{x}_{i}(t)$.

\item \textbf{Collective-instinctive component of the movement operator: } a vectorial average of the individual movements is calculated based on Eq. \ref{movcolinst}:

\begin{equation}
\label{movcolinst}
\textbf{I}=\frac{\sum^{N}_{i=1} \Delta \textbf{x}_{i} \Delta f_{i}}{\sum^{N}_{i=1} \Delta f_{i}}.
\end{equation}

Vector $\textbf{I}$ represents the weighted average of the displacements of every fish. It encompasses all fishes' move. After the vector $\textbf{I}$ computation step, every fish will move according to Eq. \ref{movinst}:

\begin{equation}
\label{movinst}
\textbf{x}_i(t+1)=\textbf{x}_{i}(t)+\textbf{I}.
\end{equation}

\item \textbf{Collective-volitive component of the movement operator: } this operator is used to automatically regulate the exploration/exploitation ability of the school during the search process. First, the barycenter $\textbf{B}$ of the school is calculated based on the position $\textbf{x}_{i}$ and the weight $W_{i}$ of each fish, as in Eq. \ref{barcalc}.

\begin{equation}
\label{barcalc}
B(t)=\frac{\sum^{N}_{i=1} x_{i}(t) W_{i}(t)}{\sum^{N}_{i=1} W_{i}(t)}.
\end{equation}

Next, if the total school weight $\sum^{N}_{i=1} W_{i}$ has increased from the last to the current iteration, the fishes are attracted to the barycenter according to Equation \ref{volAttrac}. If the total school weight has not improved, the fishes are spread away from the barycenter according to Equation \ref{volSpread}.

\begin{equation}
\label{volAttrac}
x_i(t+1)=x_{i}(t)- step_{vol} rand(0,1) * \\ \frac{x_{i}(t) - B(t)}{distance(x_{i}(t),B(t))},
\end{equation}

\begin{equation}
\label{volSpread}
x_i(t+1)=x_{i}(t)+step_{vol} rand(0,1) * \\ \frac{x_{i}(t) - B(t)}{distance(x_{i}(t),B(t))},
\end{equation}

where $step_{vol}$ defines the size of the maximum displacement performed with the use of this operator. And $distance(\textbf{x}_{i}(t),\textbf{B}(t))$ is the euclidean distance between fish $i$ position and the school barycenter. $rand(0,1)$ is a uniformly distributed random number varying from 0 up to 1.

\end{enumerate}
The parameters $step_{ind}$ and $step_{vol}$ decay linearly according to:

\begin{equation}
step_{ind}(t+1)=step_{ind}(t)-\frac{step_{ind}(initial)}{It_{max}},
\end{equation}
and similarly:
\begin{equation}
step_{vol}(t+1)=step_{vol}(t)-\frac{step_{vol}(initial)}{It_{max}},
\end{equation}
where $step_{ind}(initial)$ and $step_{vol}(initial)$ are user defined initial values for $step_{ind}$ and $step_{vol}$, respectively. $It_{max}$ is the maximum number of iterations allowed in the search process.

Besides the movement operators, FSS possesses a feeding operator used to update the weights of every fish. $W_i$ which represents the instantaneous success indicators of individual candidates. $W_i$ is updated according to Eq. \ref{feedingop}.

\begin{equation}
\label{feedingop}
W_{i}(t+1)=W_{i}(t)+\frac{\Delta f_i}{max(| \Delta f_i |)},
\end{equation}
where $W_{i}(t)$ is the weight parameter for fish $i$ in the timestamp $t$, $\Delta f_i$ is the fitness variation between the last and the new position, and $max(| \Delta f_i |)$ represents the maximum absolute value of the fitness variation among all the fishes in the school. $W$ is only allowed to vary from $1$ up to $W_{scale}$, which is a user defined attribute. The weights of all fishes are initialized with the value $W_{scale}/2$.

The pseudo-code for FSS is the following:\\

\begin{algorithmic}[1]
\label{FSSPseudoCode}
\STATE Initialize user parameters 
\STATE Initialize fishes positions randomly 
\WHILE{Stopping condition is not met}
\STATE Calculate fitness for each fish 	
\STATE Run individual operator movement
\STATE Calculate fitness for each fish 
\STATE Run feed operator 
\STATE Run collective-instinctive movement operator
\STATE Run collective-volitive movement operator
\ENDWHILE	
\end{algorithmic}

\subsection{FSS - Stagnation Avoidance Routine}
\label{FSSSAR}

As mentioned before, a modification was necessary in the original FSS aiming to make it improve its exploration ability and avoid stagnation in certain challenging search spaces. In the original version of the algorithm, the individual movement component is only allowed to move a fish if it improves the fitness. However, in a very smooth search space, there would be many moving trials with no success and the algorithm could fail to converge.

Further, the collective-volitive movement was designed to regulate the exploration/exploitation ability of the algorithm along the search process, however, in order to do so, this behavior depends on the possibility of the total weight of the school to reduce. If it does not happen, only Equation \ref{volAttrac} will be utilized in this operator. This means that the ability of attracting the fishes to the school barycenter to exploit the search space will always predominate with relation to the ability of spreading the school away from the school barycenter to allow exploration.

To solve these issues, a parameter $\alpha$ for which $0 \leq \alpha \leq 1$ was introduced in the individual component of the movement. $\alpha$ decays exponentially along with the iterations and measures a probability of a worsening allowance for each fish. It means that, every time a fish tries to move to a position that does not improve its fitness, a random number is chosen and if it is smaller than $\alpha$ the movement is allowed. Therefore, only the fishes which presented improvement in their fitnesses within the individual component of the movement can contribute to the $\textbf{I}$ vector calculation used in the collective instinctive movement. In this case, $I$ will be calculated according to Eq. \ref{isar}.

\begin{equation}
\label{isar}
\textbf{I}=\frac{\sum_{i\in N} \Delta \textbf{x}_{i} \Delta f_{i}}{\sum_{i \in N} \Delta f_{i}},
\end{equation}
where $N$ is the set of all the fishes which improved their fitnesses in the last individual movement performed.

This modification is intended to improve the algorithm exploration ability by allowing stochastic worsening movements. It is important to notice that the parameter $\alpha$ decays exponentially along the iterations and this makes the anti-stagnation routine effective only in the beginning of the search process (when decisive actions are more necessary) and irrelevant in the final of the search.

\section{MMWALBP with FSS}
\label{MMWALBPFSS}

Similarly to SALBP, the definition of the MMWALBP originates 4 different problem versions: MMWALBP-1 is the problem of minimizing the \textit{number of open workplaces} (operators) for a given \textit{cycle time}; MMWALBP-2, in which one minimizes the \textit{cycle time} for a given \textit{number of open workplaces}; MMWALBP-E is the multi-objective version which tries to optimize simultaneously the \textit{cycle time} and the \textit{number of open workplaces}; MMWALBP-F is the feasibility problem of deciding whether a given pair \textit{cycle time}/\textit{number of open workplaces} is feasible.

In this work, although four versions of the problem were conceived, only Type-1 was solved. The main issues regarding the solution of this family of problems using FSS algorithm are related to solution representation scheme and treatment of capacity and precedence constraints. These issues were tackled according to the procedures presented in the work of Hamta \textit{et al.} \cite{Hamta2013}.

\subsection{Solution representation}

\begin{figure}[htpb]
\centering
\caption[Random-keys mapping]{Random-keys mapping}
\label{randomkeys}
\includegraphics[width=6cm,height=4cm]{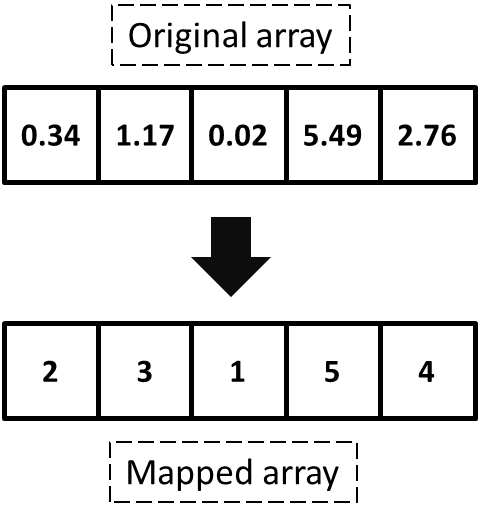}
\end{figure}

The task-oriented representation was chosen, which means that a solution is represented as an array with the size equivalent to the number of assembly tasks to be allocated. Every array position contains a task index. However, in FSS search procedure, a fish position is an array of real numbers. For this reason, a variation of the random-keys \cite{Hamta2013} procedure was utilized to map a fish position vector into a tasks indexes vector. This procedure maps the smallest value in the fish position array into number $1$, the second smallest value into number $2$, and this is repeated up to when all the array values are mapped into a task index. Figure \ref{randomkeys} shows an example of an application of the random-keys procedure on a real valued array with five dimensions.

\subsection{Precedence constraints control}

After the mapping procedure is applied, the resultant vector may contain precedence problems. In order to control that issue and to rearrange the tasks in a sequence that respects all the precedence relations, we applied the correction procedure from the work of Hamta \textit{et al}. \cite{Hamta2013}.

It is necessary to build the complete precedence matrix to apply the aforementioned procedure. The original precedence matrix registers only the direct precedence relations. Instead, the complete precedence matrix, normally referred as $M$, takes into account the indirect relations. A simple procedure can be applied to build $M$ based on the original precedence matrix. It is detailed in the following pseudocode: \\

\newpage

\begin{algorithmic}[1]
\label{buildM}
\FOR{$i \gets 1$ \TO number of tasks}
\FOR{$j \gets 1$ \TO number of tasks}
\IF{task $i$ precedes task $j$}
\STATE $M_{ij}=1$
\ELSE
\STATE $M_{ij}=0$
\ENDIF
\ENDFOR
\ENDFOR
\FOR{$i \gets 1$ \TO number of tasks}
\FOR{$j \gets 1$ \TO number of tasks}
\IF{$M_{ij}=1$}
\FOR{$K \gets 1$ \TO number of tasks}
\STATE $M_{kj} \gets M_{kj}+M_{ki}$
\IF{$M_{kj}=2$}
\STATE $M_{kj}=1$
\ENDIF
\ENDFOR			
\ENDIF
\ENDFOR
\ENDFOR
\FOR{$j \gets 1$ \TO number of tasks}
\STATE $count \gets 0$
\FOR{$i \gets 1$ \TO number of tasks}
\IF{$M_{ij}=1$}
\STATE $count \gets count +1$
\ENDIF
\ENDFOR
\STATE $M_{numberOfTasks+1,j} \gets count$
\ENDFOR
\RETURN $M$
\end{algorithmic}

With $M$ in hands, the following procedure can be used  to guarantee the correctness of a mapped vector regarding the precedence relations:

\begin{algorithmic}[1]
\label{precedenceCorrection}
\STATE \textbf{Input:} tasks array to be corrected
\STATE $Mp \gets M$ \COMMENT{copy M}
\STATE $j\gets 0$
\STATE $i \gets 1$
\STATE $n \gets number of tasks$
\WHILE{$j < n$}
\STATE $a \gets S(i)$
\IF{$Mp_{n+1,a}=0$ \AND a is not in array PS}
\STATE $j \gets j+1$
\STATE $PS[j] \gets a$
\FOR{$k \gets 1$ \TO $n$}
\IF{$Mp_{a,k}=1$}
\STATE $Mp_{a,k} \gets 0$
\STATE $Mp_{n+1,k} \gets Mp_{n+1,k}-1$
\ENDIF
\ENDFOR
\ENDIF
\STATE $i \gets i+1$
\IF{$i>n$}
\STATE $i \gets 1$
\ENDIF
\ENDWHILE
\RETURN $PS$ \COMMENT{$PS$ is the corrected array corresponding to $S$}
\end{algorithmic}

\subsection{Objective function}
\label{obfunc}

SALBP-1 utilizes the number of workstations as objective function. This implies that every fitness calculation will return a limited and discrete set of values, making the search space to exhibit many plateaus. This fact could represent a disadvantage for the use of FSS in this problem, since FSS utilizes improvement information to guide the search process (as opposed to other metaheuristics that utilize local information). This mainly impacts the individual component of the movement computation. Once it only allows fishes to move when fitness improve, if some fish is on a vast plateau in the search space, it will get trapped there. In order to tackle this issue, the objective function of SALBP-1 was changed similarly as in the work of Albuquerque \textit{et al.} \cite{albuquerque2016}. Thus, instead of the common use of the number of open workstations, the objective function selected is shown in Eq. \ref{objectiveFunction}.

\begin{equation}
\label{objectiveFunction}
minimize \left(m \times \sqrt{\sum_{k=1}^{m}(C-t_{k})^2} \right),
\end{equation}
where $m$ is the number of active workstations, $C$ is the cycle time and $t_{k}$ is the workload at station $k$. This modified objective function simultaneously improves the number of workstations and smoothness of the line balancing. Furthermore, it changes the search space allowing more variation in the fitness when fishes move.

The same objective function was utilized here to tackle MMWALBP-1 with $m$ representing the number of active workplaces (number of operators) and $t_{k}$ being the workload within the opened workplaces $k$ without taking idle times into account.

Moreover, the total workload of the assembly line $\sum_{k=1}^{m}t_{k}$ was used as a second decision criterion. If some solution has the same fitness value as the other, the considered solution will be the one with the lower total workload. This decision was made to make the algorithm to give preference to the solutions which cause the lowest number of displacements between different work zones within the workstations.

\subsection{General solution procedure}
\label{generalSolution}

\begin{figure}[htpb]
\centering
\caption[Solution procedure]{Solution procedure}
\label{solutionProcedure}
\includegraphics[width=8cm,height=5cm]{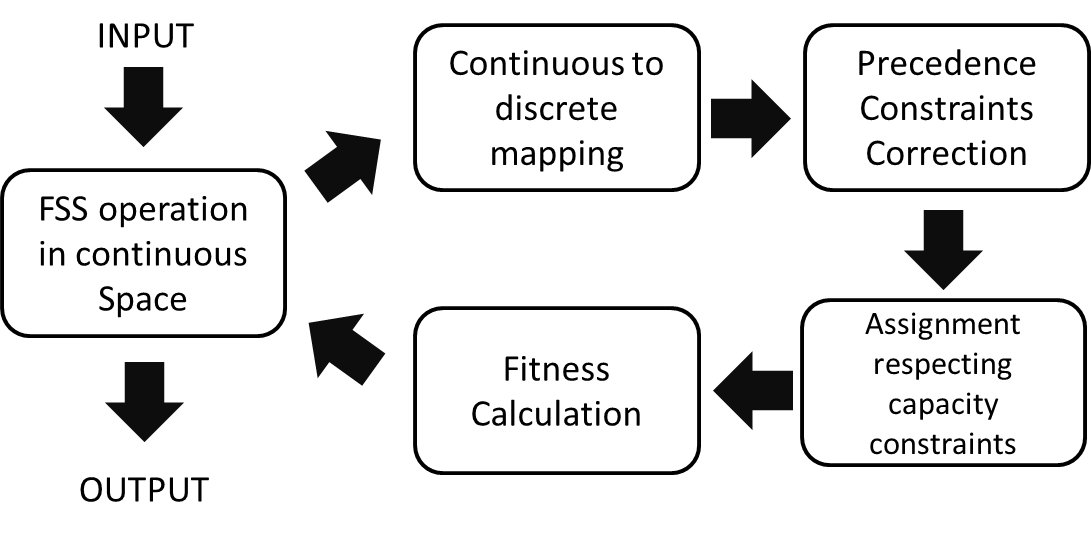}
\end{figure}

The approach used to solve MMWALBP-1 with FSS algorithm does not affect its heuristic search structure showed in the pseudo code \ref{FSSPseudoCode}. As can be seen in Figure \ref{solutionProcedure}, the algorithm runs its original procedure and all the issues regarding the balancing problem are taken into account when the search procedure demands the fitness calculation for every fish.

The assignment of tasks into workplaces used in this work is a variation of the procedure presented in the work of Scholl \cite{scholl1999balancing} for the task oriented approach.

The mentioned procedure consists basically on iteratively select a single available task and assign it to a workstation to which it can be assigned. According to Scholl and Becker \cite{Scholl2006} a task $j$ is assignable if:

\begin{itemize}

\item all the precedent tasks of $j$ are already assigned;
\item the total time $t_k$ allocated in workstation $k$, including the time of task $j$, does not exceed the cycle time;

\end{itemize}

\begin{figure}[htpb]
\centering
\caption[Assignment procedure]{Assignment procedure}
\label{assignmentProcedure}
\includegraphics[width=13cm,height=18cm]{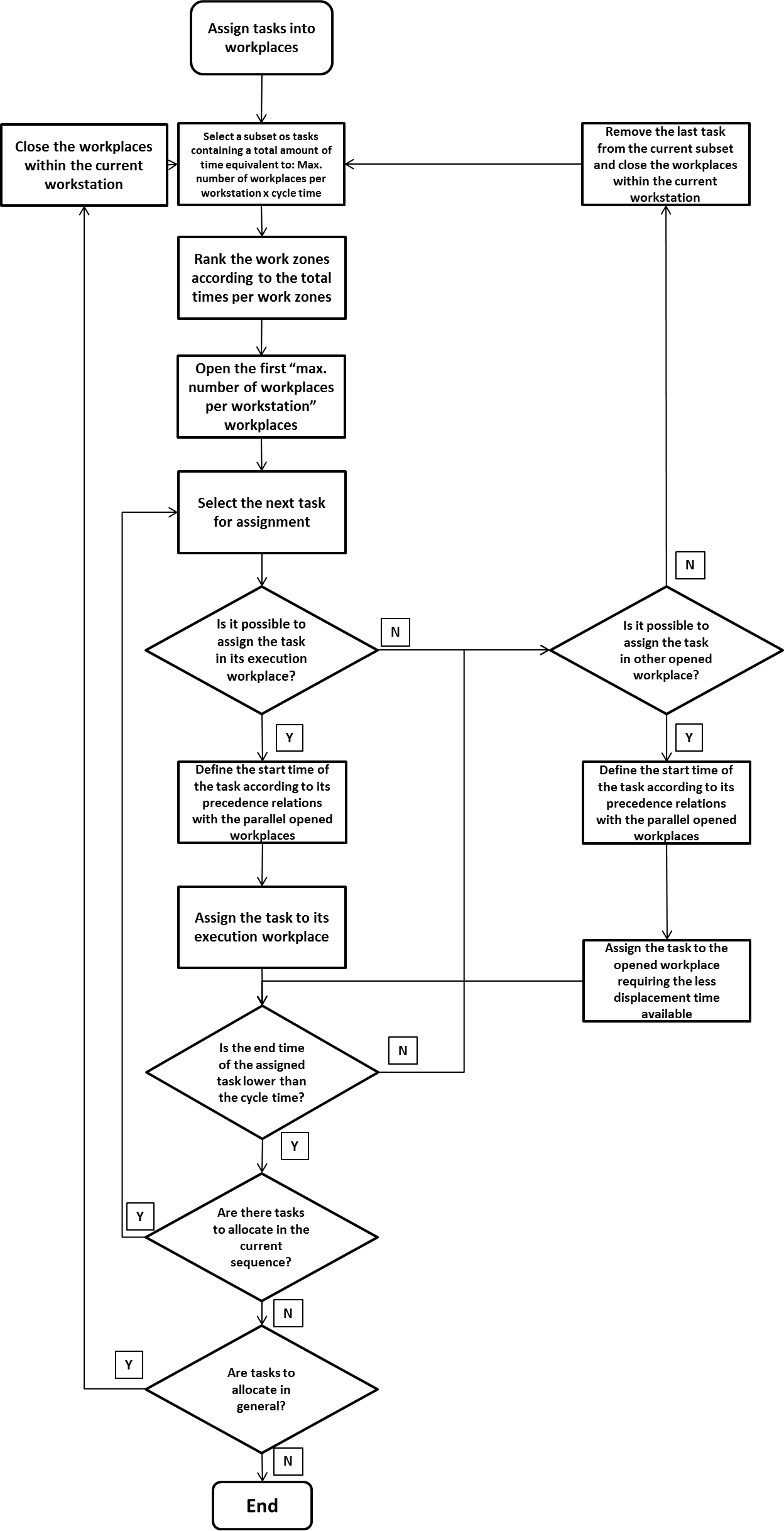}
\end{figure}

A variation of the aforementioned procedure was conceived to convert a task sequence array in a feasible assignment for MMWALBP-1. This procedure is detailed in Figure \ref{assignmentProcedure} and its main steps are:

\begin{enumerate}

\item Select the largest possible subset of tasks with total time lower than: \textit{maximum number of opened workplaces per workstation} $\times$ \textit{cycle time};
\item Rank the work zones based on the total time (each task has an attribute indicating where it should be performed);
\item Open the \textit{maximum number of active workplaces per workstation} (user defined parameter) highest ranked work zones;
\item Iteratively assign the tasks: the first assignment trial must check if there is an opened workplace corresponding to the work zone in which the task should be performed. If it is not the case, the assignment attempts should be carried out prioritizing the open workplaces that require the least amount of displacement time;
\item Once a task is assigned, its time must be corrected including the displacement time (if this is the case). The start/end times must be updated taking into account the precedence relations in the tasks assigned in all the active workplaces within the same workstation.

\end{enumerate}

\section{Experiments and Tests}
\label{experiments}

\subsection{Test Data Definition}
\label{testsets}

MMWALBP was tackled for the first time in this work, hence there is no data set available to compare the solution procedure utilized with others techniques. The test data utilized for evaluating the procedure proposed was derived from test instances of SALBP-1. 

All test instances used in this work were taken from the assembly line data repository \url{www.assembly-line-balancing.de}. Three problems were chosen with different number of tasks to assign. These problems are referred as $n=20\_26$ (small), $n=50\_25$ (medium) and $n=100\_34$ (large).

As mentioned before, the tests chosen must be converted from SALBP-1 instances to MMWALBP-1 ones. In order to do so, it was necessary to define the production plans. Two variations were chosen with the first one containing $4$ different models and a production level of $200$ models and the second one being composed of $50$ models and a production level of $998$ products.

Moreover, the definition of the work zone in which each task should be performed had to be set. We made it by sorting random integer values in $\{[0; 8]-[4]\}$. The value $4$ indicates the work zone in the inside of the product. It was chosen not to be used in this work.

Furthermore, a binary matrix with dimensions \textit{number of tasks} $\times$ \textit{number of models} was defined with its values chosen randomly with equal probabilities. This matrix is intended to define in which models each task must be performed. Table \ref{datasets} indicates the features of each of the data sets defined.

\begin{table}[h!]
\centering
\caption{Data sets features}
\label{datasets}
\resizebox{12cm}{!}{\begin{tabular}{cccccc}
\textit{Dataset ID}        & \textit{Number of tasks} & \textit{Number of Models} & \textit{Cycle time} & \textit{Workload} & \textit{Workload / cycle time} \\ \hline
\textbf{$n=20\_26\_4M$}   & 20                       & 4                         & 1000                & 5144.22           & 5.14                           \\
\textbf{$n=20\_26\_50M$}  & 20                       & 50                        & 1000                & 5342.60           & 5.34                           \\
\textbf{$n=50\_25\_4M$}   & 50                       & 4                         & 1000                & 3395.54           & 3.40                           \\
\textbf{$n=50\_25\_50M$}  & 50                       & 50                        & 1000                & 2871.60           & 2.87                           \\
\textbf{$n=100\_34\_4M$}  & 100                      & 4                         & 1000                & 7888.34           & 7.89                           \\
\textbf{$n=100\_34\_50M$} & 100                      & 50                        & 1000                & 7171.42           & 7.17                           \\ \hline
\end{tabular}}
\end{table}

The matrix containing time increments to allow times correction when some task require a displacement from a work zone to other one within the same workstation was set as detailed in Table \ref{timesmatrix2}.

\begin{table}[h!]
\centering
\caption{Times correction matrix}
\label{timesmatrix2}
\begin{tabular}{cccccccccc}
& \textit{\textbf{0}} & \textit{\textbf{1}} & \textit{\textbf{2}} & \textit{\textbf{3}} & \textit{\textbf{4}} & \textit{\textbf{5}} & \textit{\textbf{6}} & \textit{\textbf{7}} & \textit{\textbf{8}} \\
\textit{\textbf{0}} & 0                   & 54                  & 27                  & 27                  & 0                   & 13.5                & 13.5                & 40.5                & 40.5                \\
\textit{\textbf{1}} & 54                  & 0                   & 27                  & 27                  & 0                   & 40.5                & 40.5                & 13.5                & 13.5                \\
\textit{\textbf{2}} & 27                  & 27                  & 0                   & 54                  & 0                   & 13.5                & 40.5                & 13.5                & 40.5                \\
\textit{\textbf{3}} & 27                  & 27                  & 54                  & 0                   & 0                   & 40.5                & 13.5                & 40.5                & 13.5                \\
\textit{\textbf{4}} & 0                   & 0                   & 0                   & 0                   & 0                   & 0                   & 0                   & 0                   & 0                   \\
\textit{\textbf{5}} & 13.5                & 40.5                & 13.5                & 40.5                & 0                   & 0                   & 27                  & 27                  & 54                  \\
\textit{\textbf{6}} & 13.5                & 40.5                & 40.5                & 13.5                & 0                   & 27                  & 0                   & 54                  & 27                  \\
\textit{\textbf{7}} & 40.5                & 13.5                & 13.5                & 40.5                & 0                   & 27                  & 54                  & 0                   & 27                  \\
\textit{\textbf{8}} & 40.5                & 13.5                & 40.5                & 13.5                & 0                   & 54                  & 27                  & 27                  & 0                  
\end{tabular}
\end{table}

\subsection{Parameters definition}
\label{parameters}

Some parameters should be defined in order for the Fish School Search algorithm to solve MMWALBP-1. However, there is not any reference in literature indicating the best values for this specific application. Therefore, the same procedure utilized in the work of Albuquerque \textit{et al.} \cite{albuquerque2016} was applied. It means that three parameters, $Step_{ind}$, $Step_{vol}$ and $W_{scale}$ had two values defined, being the smallest of these values labelled as \textit{low level} and the highest as \textit{high level} resulting in $2^3 = 8$ different combinations of values of the three selected parameters. These three parameter affect the exploratory ability of FSS, and tuning them is a relevant issue to be solved to guarantee convergence to good results. High and low levels for each parameters were chosen based on the values used in the work of Monteiro \textit{et al.} \cite{Filho2008}. Table \ref{paramValues} summarizes the values used for each variable.
\begin{table}[h!]
\centering
\caption{High and low values used for each of the parameters}
\label{paramValues}
\begin{tabular}{c|ccc}
\hline
Level & $W_{scale}$ & $Step_{ind}$ & $Step_{vol}$ \\ \hline
High  & 10000       & 20           & 20            \\
Low   & 1000        & 2            & 0.2            \\ \hline
\end{tabular}
\end{table}

Data set $n=100\_34\_4M$ was chosen for this step of the work. As this is the data set with highest number of tasks, it is assumed that a good set of parameter for this problem will also generate good results in smaller data sets.

Table \ref{Levels} summarizes the test configurations executed in this study. In this table, each different configuration has an ID and the FSS original version (vanilla version) is refered as FSS-V. Each one of the test configurations was executed $30$ times. All tests performed within this work occurred within $\left[-100; 100\right]$.

\begin{table}[h!]
\centering
\caption{Configurations of the tests executed}
\label{Levels}
\begin{tabular}{c|ccccc}
\hline
ID & Version & $W_{scale}$ & $Step_{ind}$ & $Step_{vol}$ & Problem \\ \hline
1 & FSS-V   & Low    & Low     & Low     & Large   \\
2 & FSS-V   & High   & Low     & Low     & Large   \\
3 & FSS-V   & Low    & High    & Low     & Large   \\
4 & FSS-V   & High   & High    & Low     & Large   \\
5 & FSS-V   & Low    & Low     & High    & Large   \\
6 & FSS-V   & High   & Low     & High    & Large   \\
7 & FSS-V   & Low    & High    & High    & Large   \\
8 & FSS-V   & High   & High    & High    & Large   \\
9 & FSS-SAR & Low    & Low     & Low     & Large   \\
10 & FSS-SAR & High   & Low     & Low     & Large   \\
11 & FSS-SAR & Low    & High    & Low     & Large   \\
12 & FSS-SAR & High   & High    & Low     & Large   \\
13 & FSS-SAR & Low    & Low     & High    & Large   \\
14 & FSS-SAR & High   & Low     & High    & Large   \\
15 & FSS-SAR & Low    & High    & High    & Large   \\
16 & FSS-SAR & High   & High    & High    & Large   \\ \hline
\end{tabular}
\end{table}

Table \ref{parSelection} shows the results found for $500$ iterations and $30$ runs of each combination of FSS version and set of parameters. The $\alpha$ parameter in the SAR version of FSS was chosen to decay according to $\alpha=0.8e^{-0.007t}$ (original value as suggested by the original proponents). $t$ is the current iteration number. The maximum number of open workplaces per workstation, a user defined attribute which should be within $[1;8]$, was set to the intermediary value $3$ in all the tests performed.

\begin{table}[h!]
\centering
\caption{Results for parameter selection}
\label{parSelection}
\resizebox{12cm}{!}{
\begin{tabular}{cccccccccc}
\multicolumn{1}{l}{} & \multicolumn{1}{l}{}                     & \multicolumn{4}{c}{\textit{\textbf{Objective Function}}}                              & \multicolumn{4}{c}{\textit{\textbf{Opened Workplaces}}}          \\ \hline
\textbf{Version}     & \multicolumn{1}{c|}{\textbf{Param. Set}} & \textbf{Mean} & \textbf{Std. Dev.} & \textbf{Max} & \multicolumn{1}{c|}{\textbf{Min}} & \textbf{Mean} & \textbf{Std. Dev.} & \textbf{Max} & \textbf{Min} \\ \hline
FSS-V                & \multicolumn{1}{c|}{1}                   & 104082.75     & 24.14              & 104130.75    & \multicolumn{1}{c|}{104009.88}    & 11            & 0                  & 11           & 11           \\
FSS-V                & \multicolumn{1}{c|}{2}                   & 104086.87     & 23.64              & 104130.97    & \multicolumn{1}{c|}{104038.49}    & 11            & 0                  & 11           & 11           \\
FSS-V                & \multicolumn{1}{c|}{3}                   & 104074.78     & 25.65              & 104131.42    & \multicolumn{1}{c|}{104028.34}    & 11            & 0                  & 11           & 11           \\
FSS-V                & \multicolumn{1}{c|}{4}                   & 104075.41     & 25.2               & 104118.25    & \multicolumn{1}{c|}{104011.96}    & 11            & 0                  & 11           & 11           \\
FSS-V                & \multicolumn{1}{c|}{5}                   & 104078.12     & 20.68              & 104115.63    & \multicolumn{1}{c|}{104041.45}    & 11            & 0                  & 11           & 11           \\
FSS-V                & \multicolumn{1}{c|}{6}                   & 104087.09     & 26.69              & 104132.13    & \multicolumn{1}{c|}{104026.65}    & 11            & 0                  & 11           & 11           \\
FSS-V                & \multicolumn{1}{c|}{7}                   & 104081.52     & 28.61              & 104139.19    & \multicolumn{1}{c|}{104016.93}    & 11            & 0                  & 11           & 11           \\
FSS-V                & \multicolumn{1}{c|}{8}                   & 104075.17     & 32.12              & 104175.33    & \multicolumn{1}{c|}{104003.71}    & 11            & 0                  & 11           & 11           \\
FSS-SAR              & \multicolumn{1}{c|}{1}                   & 104077.24     & 28.97              & 104138.57    & \multicolumn{1}{c|}{104017.75}    & 11            & 0                  & 11           & 11           \\
FSS-SAR              & \multicolumn{1}{c|}{2}                   & 104080.03     & 31.04              & 104149.65    & \multicolumn{1}{c|}{104012.1}     & 11            & 0                  & 11           & 11           \\
FSS-SAR              & \multicolumn{1}{c|}{3}                   & 104072.83     & 20.61              & 104108.71    & \multicolumn{1}{c|}{104017.93}    & 11            & 0                  & 11           & 11           \\
FSS-SAR              & \multicolumn{1}{c|}{4}                   & 104084.92     & 23.72              & 104137.17    & \multicolumn{1}{c|}{104036.74}    & 11            & 0                  & 11           & 11           \\
FSS-SAR              & \multicolumn{1}{c|}{5}                   & 104092.13     & 27.31              & 104145.54    & \multicolumn{1}{c|}{104043.22}    & 11            & 0                  & 11           & 11           \\
FSS-SAR              & \multicolumn{1}{c|}{6}                   & 104084.81     & 29.68              & 104139.54    & \multicolumn{1}{c|}{104006.47}    & 11            & 0                  & 11           & 11           \\
FSS-SAR              & \multicolumn{1}{c|}{7}                   & 103772.12     & 1655.26            & 104120.81    & \multicolumn{1}{c|}{95008.94}     & 10.97         & 0.18               & 11           & 10           \\
FSS-SAR              & \multicolumn{1}{c|}{8}                   & 104067.44     & 26.63              & 104112.32    & \multicolumn{1}{c|}{104002.26}    & 11            & 0                  & 11           & 11           \\ \hline
\end{tabular}}
\end{table}

From the results, it can be seen that the experiment corresponding to ID 15 was the only able to reach the value of 10 open workplaces. Hence, the corresponding set of parameters was chosen for the experiments in this work.

\subsection{Results}
\label{results}

In order to evaluate the performance of FSS and FSS-SAR in solving MMWALBP-1, we compared the results obtained with MMWALBP-1 solution using PSO. In the PSO version chosen \cite{clerc2002particle}, for a particle $i$, its position in iteration $t+1$ is defined in Eq. \ref{psoeq}:

\begin{equation}
\label{psoeq}
\textit{x}_i(t+1) = \textit{x}_i(t) + \chi [\textit{v}_{i} + c_1r_1(\textit{pb}_{i}-\textit{x}_{i}(t)) + \\ c_2r_{2}(\textit{gb}_{i} - \textit{x}_{i}(t))],	\end{equation}
where $\chi = \frac{2}{|2 - (c_1 + c_2) - \sqrt{(c_1 + c_2)((c_1 + c_2) - 4)}|}$ is known as constriction factor and $r_1$ and $r_2$ are uniformly distributed random numbers in the interval $[0; 1]$. In this version $c_1$ and $c_2$ must satisfy $c_1 + c_2 \geq 4$. For this work we have chosen $c_1 = c_2 = 2.1$. The solution of MMWALP-1 with PSO follows the same flow for FSS, as described in Section \ref{generalSolution}. 

FSS, FSS-SAR and PSO were used to solve all the six different test instances defined in Section \ref{testsets}. We solved the problem instances using the parameters defined within the procedure presented in Section \ref{parameters} and each test case was repeated $450$ times.

The widely used Analysis of Variance (ANOVA) technique \cite{montgomery2010applied} was used to establish whether the results obtained are significantly different and, if so, which algorithm is better for each one of the criteria considered.

To apply ANOVA, normality of data should be guaranteed. Thus, the $450$ results for each simulation were grouped in samples of size $15$ and the means of those were considered as input data for the ANOVA, which results in $30$ samples per algorithm for each instance. We applied Shapiro-Wilk test \cite{razali2011power} and then concluded that normality is guaranteed for the results obtained in all test cases.

Only the workload and smoothness of the solutions generated by each technique were analyzed because all algorithms were able to reach the same values of open workplaces as can be seen in Table \ref{openedWorkplaces}. The lower bounds for each data set is shown in Table \ref{datasets} (\textit{Workload / cycle time}). In the case of the small data set with $4$ models, one single solution within $450$ trials solved by FSS-V generated the output $10$ instead of 11 such as FSS-SAR and PSO.

\begin{table}[h!]
\centering
\caption{Best number of open workplaces achieved in each data set}
\label{openedWorkplaces}
\resizebox{\textwidth}{!}{
\begin{tabular}{c|cccccc}
\textbf{Algorithm} & \textbf{Small - 4M} & \textbf{Small - 50M} & \textbf{Medium - 4M} & \textbf{Medium - 50M} & \textbf{Large - 4M} & \textbf{Large - 50M} \\ \hline
\textbf{FSS-V}     & 6                   & 6                    & 6                    & 6                     & 10                  & 9                    \\
\textbf{FSS-SAR}   & 6                   & 6                    & 6                    & 6                     & 11                  & 9                    \\
\textbf{PSO}       & 6                   & 6                    & 6                    & 6                     & 11                  & 9                    \\ \hline
\end{tabular}}
\end{table}

For all test instances, an one-way ANOVA with $95 \%$ of confidence was performed. Calculated degrees of freedom were $v_1 = 2$ and $v_2 = 87$, thus $F_{ref} = 4.89$. The values of $F_{calculated}$ for each ANOVA are shown in Table \ref{ANOVARes}.

\begin{table}[h!]
\centering
\caption{$F_{calculated}$ for the results generated with the different algorithms}
\label{ANOVARes}
\resizebox{\textwidth}{!}{
\begin{tabular}{c|cccccc}
\hline
Criterion  & Small - 4M & Small - 50M & Medium - 4M & Medium - 50M & Large - 4M & Large - 50M \\ \hline
Smoothness & 128.50     & 7.87        & 541.87      & 947.70       & 34.39      & \textbf{1.36}        \\
Workload   & 257.70     & 84.61       & 325.08      & 532.98       & \textbf{3.48}       & \textbf{1.95}        \\ \hline
\end{tabular}}
\end{table}

From Table \ref{ANOVARes}, it can be seen that some of the values of $F_{calculated}$ are lower than $F_{ref}$ which means that we can not conclude that there is performance difference in those cases.

\begin{figure}[H]
\centering
\caption{Confidence intervals based on the pooled standard deviation for the small data set with 4 models}
\subfigure[ref1][Smoothness - 4 models]{\includegraphics[width=5cm]{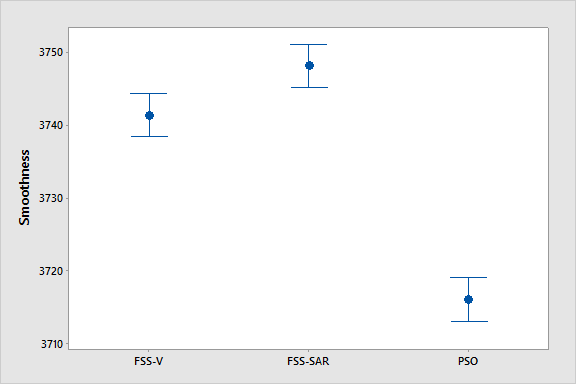}}
\qquad
\subfigure[ref2][Workload - 4 models]{\includegraphics[width=5cm]{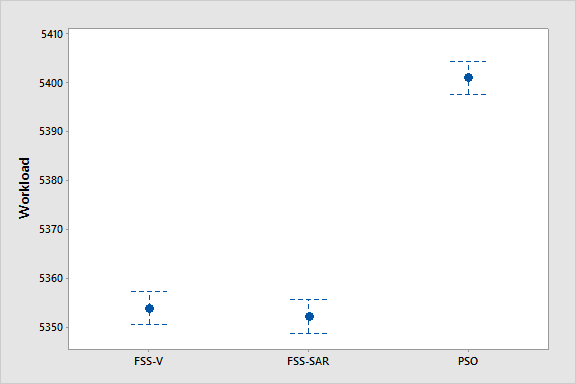}}
\label{small1}
\end{figure}

\begin{figure}[H]
\centering
\caption{Confidence intervals based on the pooled standard deviation for the small data set with 50 models}
\subfigure[ref1][Smoothness - 50 models]{\includegraphics[width=5cm]{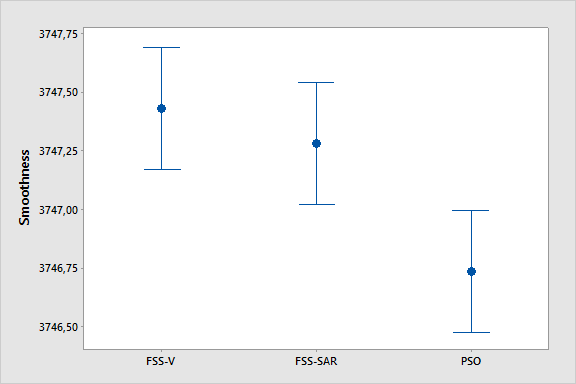}}
\qquad
\subfigure[ref2][Workload - 50 models]{\includegraphics[width=5cm]{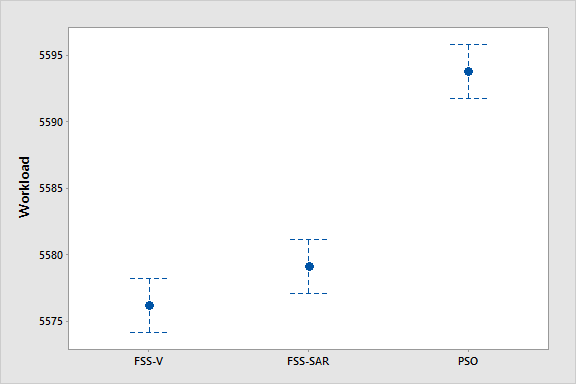}}
\label{small2}
\end{figure}

\begin{figure}[H]
\centering
\caption{Confidence intervals based on the pooled standard deviation for the medium data set with 4 models}
\subfigure[ref1][Smoothness - 4 models]{\includegraphics[width=5cm]{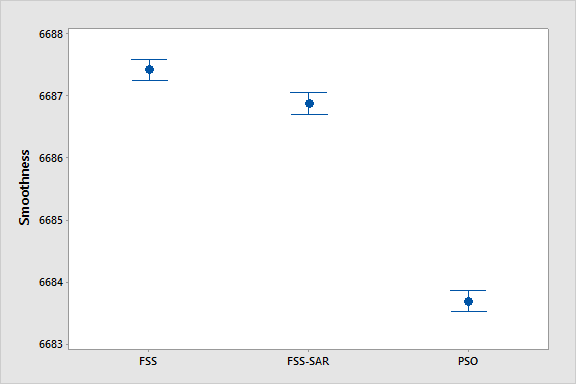}}
\qquad
\subfigure[ref2][Workload - 4 models]{\includegraphics[width=5cm]{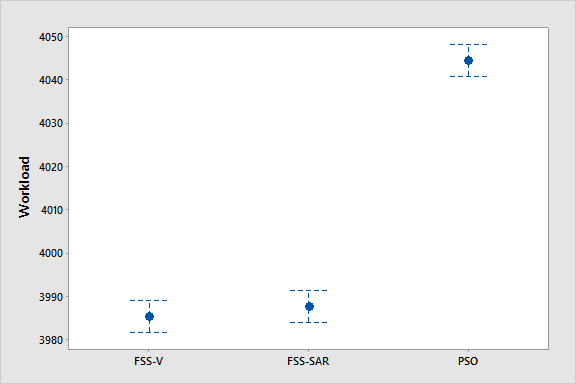}}
\label{medium1}
\end{figure}

\begin{figure}[H]
\centering
\caption{Confidence intervals based on the pooled standard deviation for the medium data set with 50 models}
\subfigure[ref1][Smoothness - 50 models]{\includegraphics[width=5cm]{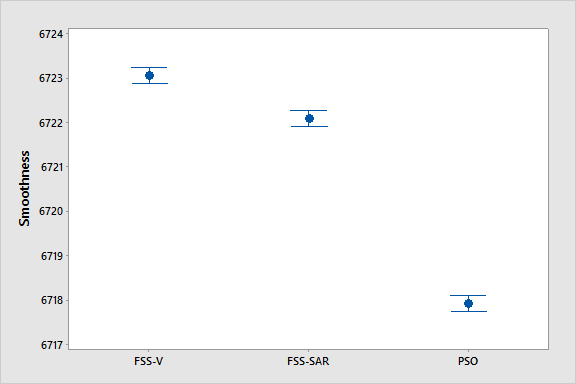}}
\qquad
\subfigure[ref2][Workload - 50 models]{\includegraphics[width=5cm]{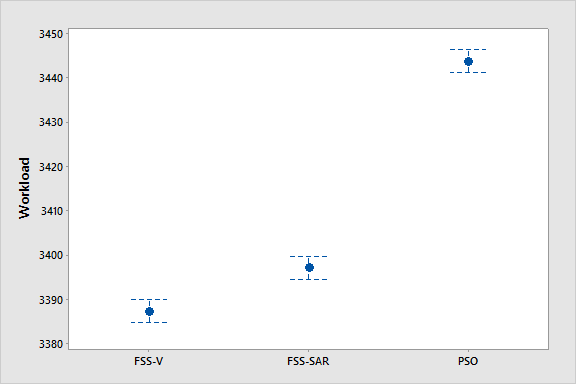}}
\label{medium2}
\end{figure}

Regarding the small and medium data sets, Figures \ref{small1}, \ref{small2}, \ref{medium1} and \ref{medium2} show the pooled confidence intervals in both 4 models and 50 models cases. It can be seen that PSO achieved the best values of smoothness but FSS-V and FSS-SAR presented better results for the workload in all cases. It could not be perceived which of the FSS versions is better in the small data sets. However, FSS-SAR presented better results than FSS-V for smoothness in the medium data set with 4 models and in both workload and smoothness for 50 models.

\begin{figure}[h]
\centering
\caption{Confidence intervals based on the pooled standard deviation for the large data set with 4 models}
\subfigure[ref1][Smoothness - 4 models]{\includegraphics[width=5cm]{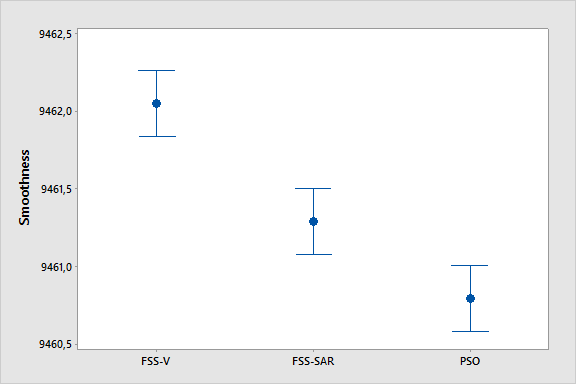}}
\qquad
\subfigure[ref2][Workload - 4 models]{\includegraphics[width=5cm]{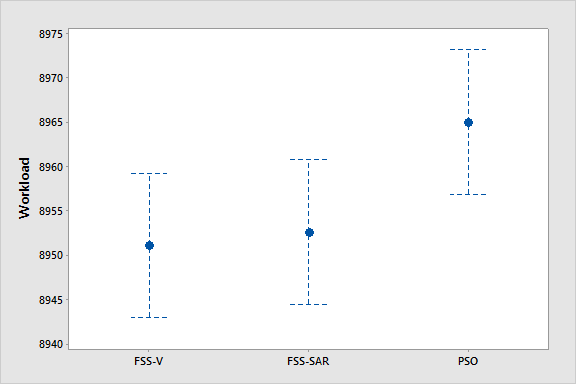}}
\label{large1}
\end{figure}

\begin{figure}[h]
\centering
\caption{Confidence intervals based on the pooled standard deviation for the large data set with 50 models}
\subfigure[ref1][Smoothness - 50 models]{\includegraphics[width=5cm]{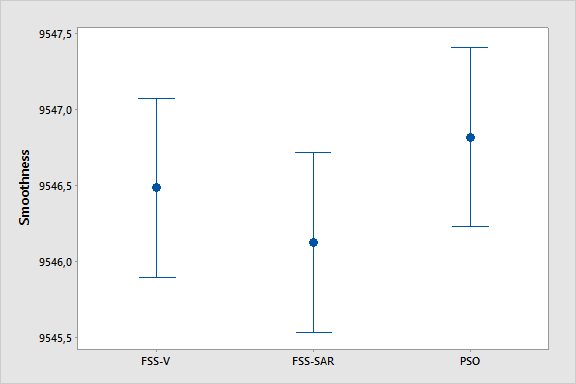}}
\qquad
\subfigure[ref2][Workload - 50 models]{\includegraphics[width=5cm]{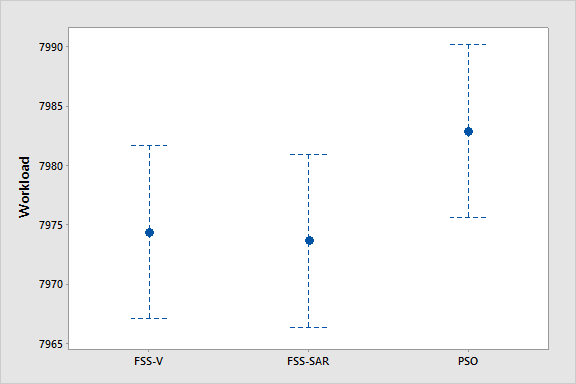}}
\label{large2}
\end{figure}

The pooled confidence intervals in the large data sets for smoothness and workload are shown in Figures \ref{large1} and \ref{large2}. 

As it was mentioned before, specifically in the large data sets, the $F$ values in Table \ref{ANOVARes} as well as the overlaps in the pooled confidence intervals do not allow us to conclude performances differences within the algorithms in these data sets. However, in the case of Large - 4M data set with 4 models we can conclude that PSO presented better smoothness than FSS versions and FSS-SAR was better than FSS-V in this criterion.

\section{Conclusion}
\label{conclusion}

In this article it was introduced a novel problem version for the Assembly Line Balancing family of problems that takes into account some features of real assembly lines, mainly lines producing big sized products such as cars and trucks. The new problem version is referred as Mixed Model Workplace Time-dependent Assembly Line Balancing Problem (MMWALBP) and got inspiration from the VWALBP proposed in the work of Becker and Scholl  \cite{Becker2009}. However, a different approach was applied as displacements between workplaces within the same workstation are now allowed.

Some different MMWALBP instances were then solved with three metaheuristic techniques: FSS-V, FSS-SAR and PSO. In order to guarantee the reliability of conclusions taken from the resulting data of experiments performed, results were compared through Analysis of Variance. Regarding the number of active workplaces (number of open workplaces) all algorithms achieved the same outputs.

However, considering other features, such as total workload and smoothness of workload among the active workplaces it can be seen that there are differences in the outputs generated from the algorithms utilized. In general, PSO presented solutions with better smoothness and FSS versions presented solutions with lower workload. It could not be perceived a clear performance distinction between FSS-V and FSS-SAR. The 3 algorithms presented indistinct results in the large data sets.

As future works, the related sequencing problem from the mixed model feature of MMWALBP could be solved using a simultaneous approach. Further, MMWALBP might become more eficient if the assignment procedure utilizes idle times to compensate displacement times, which needs experimental tests. Moreover, in this first work with MMWALBP, position constraints were not taken into account and they should be included in future assessments of this problem. Finally, to evaluate the performance variation of FSS-SAR when varying the decay mode of $\alpha$ seems a relevant issue to be tried out as well.

\bibliographystyle{abbrv}
\bibliography{bibliografia}

\end{document}